\documentclass[10pt,twocolumn,letterpaper]{article}

\usepackage[pagenumbers]{cvpr} 
\usepackage{enumitem}
\usepackage[dvipsnames]{xcolor}

\usepackage{algorithm}
\usepackage{algorithmic}
\usepackage{multirow}
\usepackage{multicol}
\usepackage{booktabs}  
\usepackage{threeparttable}
\usepackage{amsfonts}       
\usepackage{amssymb}     
\usepackage{amsmath}     
\usepackage{nicefrac}       
\usepackage{microtype}      
\usepackage{xcolor}         
\usepackage{bm}     
\usepackage{xfrac} 
\usepackage{array}
\usepackage{makecell}
\usepackage{wrapfig}
\usepackage{enumitem}

\newcommand{\PreserveBackslash}[1]{\let\temp=\\#1\let\\=\temp}
\newcolumntype{C}[1]{>{\PreserveBackslash\centering}p{#1}}
\newcolumntype{R}[1]{>{\PreserveBackslash\raggedleft}p{#1}}
\newcolumntype{L}[1]{>{\PreserveBackslash\raggedright}p{#1}}
\newcommand{\bihan}[1]{\textcolor{black}{#1}}
\definecolor{cvprblue}{rgb}{0.21,0.49,0.74}
\usepackage[pagebackref,breaklinks,colorlinks,citecolor=cvprblue]{hyperref}

\def\paperID{3918} 
\def\confName{CVPR}
\def\confYear{2024}
\def\eg{\textit{e.g.}}
\def\ie{\textit{i.e.}}

\title{SinSR: Diffusion-Based Image Super-Resolution in a Single Step}

\author{
{Yufei Wang$^{1,3\dag}$,
Wenhan Yang$^2$, 
Xinyuan Chen$^3$, 
Yaohui Wang$^3$, 
Lanqing Guo$^1$,}\\ 
{
Lap-Pui Chau$^4$, 
Ziwei Liu$^1$, 
Yu Qiao$^3$, 
Alex C. Kot$^1$, 
Bihan Wen$^1$}\\
{\fontsize{10}{12}\selectfont
$^1$Nanyang Technological University \quad $^2$Peng Cheng Laboratory} \\
{\fontsize{10}{12}\selectfont
$^3$Shanghai Artificial Intelligence Laboratory \quad $^4$The Hong Kong Polytechnic University}\\
}

\begin{document}
\maketitle
\let\thefootnote\relax\footnotetext{$^\dag$ Work done as an intern at Shanghai AI Laboratory.}
\begin{abstract}
\vspace{-0.2cm}
\bihan{While super-resolution (SR) methods based on diffusion models exhibit promising results, their practical application is hindered by the substantial number of required inference steps.}
Recent \bihan{methods utilize the degraded images in the initial state}, thereby shortening the Markov chain. 
%
\bihan{Nevertheless, these solutions either rely on a precise formulation of the degradation process or still necessitate a relatively lengthy generation path (e.g., 15 iterations).}
%
\bihan{To enhance inference speed, we propose a simple yet effective method for achieving single-step SR generation, named \textbf{SinSR}.}
%
\bihan{Specifically,}
we first derive a deterministic sampling process from the most recent state-of-the-art (SOTA) method for accelerating diffusion-based SR.
\bihan{This allows the mapping between the input random noise and the generated high-resolution image to be obtained in a reduced and acceptable number of inference steps during training.}
%
\bihan{We show} that this deterministic mapping \bihan{can be} distilled into a student model that performs SR within only one inference step.
%
\bihan{Additionally, we propose a novel consistency-preserving loss to simultaneously leverage the ground-truth image during the distillation process, ensuring that the performance of the student model is not solely} bound by the feature manifold of the teacher model, resulting in further performance improvement.
%
Extensive experiments \bihan{conducted} on synthetic and real-world datasets demonstrate that the proposed method can achieve comparable or even superior performance \bihan{compared to} both previous SOTA methods and the teacher model,
\bihan{in just one sampling step, resulting in a remarkable up to $\times 10$ speedup for inference.}
Our code will be released at {\small \url{https://github.com/wyf0912/SinSR/}}.
\end{abstract}

\vspace{-0.3cm}
\section{Introduction}
\begin{figure}[t]
    \centering
    \includegraphics[width=\linewidth]{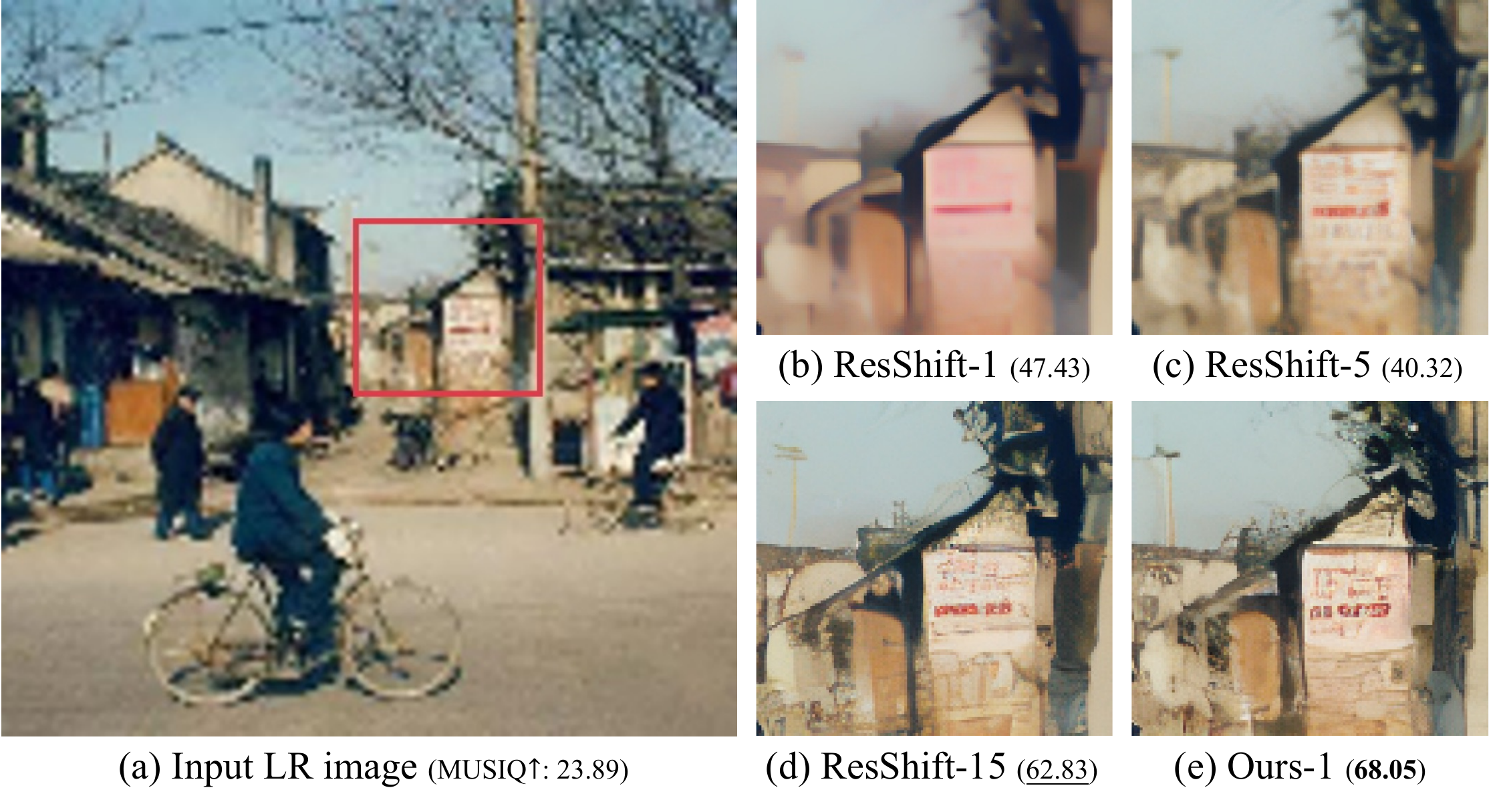}
    \vspace{-0.6cm}
    \caption{A comparison between the most recent SOTA method ResShift~\cite{yue2023resshift} for the acceleration of diffusion-based SR and the proposed method. We achieve on-par or even superior perceptual quality using only one inference step. (``-N” behind the method name represents the number of inference steps, and the value in the bracket is the quantitative result measured by MUSIQ$\uparrow$~\cite{ke2021musiq}.)}
    \label{fig:intro}
    \vspace{-0.5cm}
\end{figure}
\bihan{Image super-resolution (SR) aims to reconstruct a high-resolution image from a given low-resolution (LR) counterpart~\cite{wang2020deep}. Recently, diffusion models, known for their effectiveness in modeling complex distributions, have gained widespread adoption and demonstrated remarkable performance in SR tasks, particularly in terms of perceptual quality.}

\bihan{Specifically, current strategies for employing diffusion models can be broadly categorized into two streams:}
concatenating the LR image to the input of the denoiser in the diffusion models~\cite{saharia2022image, rombach2022high}, and adjusting the inverse process of a pre-trained diffusion model~\cite{kawar2022denoising, chung2022come, Choi_Kim_Jeong_Gwon_Yoon_2021}.
\bihan{Despite achieving promising results, both strategies encounter computational efficiency issues.}
\bihan{Notably}, the initial state of \bihan{these} conditional diffusion models is a pure Gaussian noise without using the prior knowledge from the LR image. 
\bihan{Consequently, a substantial number of inference steps are required to achieve satisfactory performance, significantly hindering the practical applications of diffusion-based SR techniques.}

Efforts have been made to enhance the sampling efficiency of diffusion models, \bihan{leading to} various techniques proposed~\cite{song2020denoising, nichol2021improved, lu2022dpm}. 
However, in the \bihan{realm} of low-level vision where \bihan{maintaining} high fidelity is critical, these techniques \bihan{often fall short} as they achieve acceleration at the cost of performance. 
More recently, innovative techniques have \bihan{emerged} to reformulate the diffusion process in image restoration tasks, focusing on improving the signal-to-noise ratio of the initial diffusion state \bihan{and thereby} shorten the Markov chain. For instance, \cite{wang2023exposurediffusion} initiates the denoising diffusion process with the input noisy image, while in the SR task, \cite{yue2023resshift} models the initial step as a combination of the LR image and random noise. 
Nonetheless, even \bihan{in} these most recent works~\cite{wang2023exposurediffusion, yue2023resshift}, limitations persist. For instance, while \cite{wang2023exposurediffusion} shows promising results within just three inference steps, it requires a clear formulation of the image degradation process. Besides, \cite{yue2023resshift} still \bihan{necessitates} $15$ inference steps and exhibits degraded performance with \bihan{noticeable} artifacts if the number of inference steps is further reduced.

\begin{figure}
    \centering
    \includegraphics[width=\linewidth]{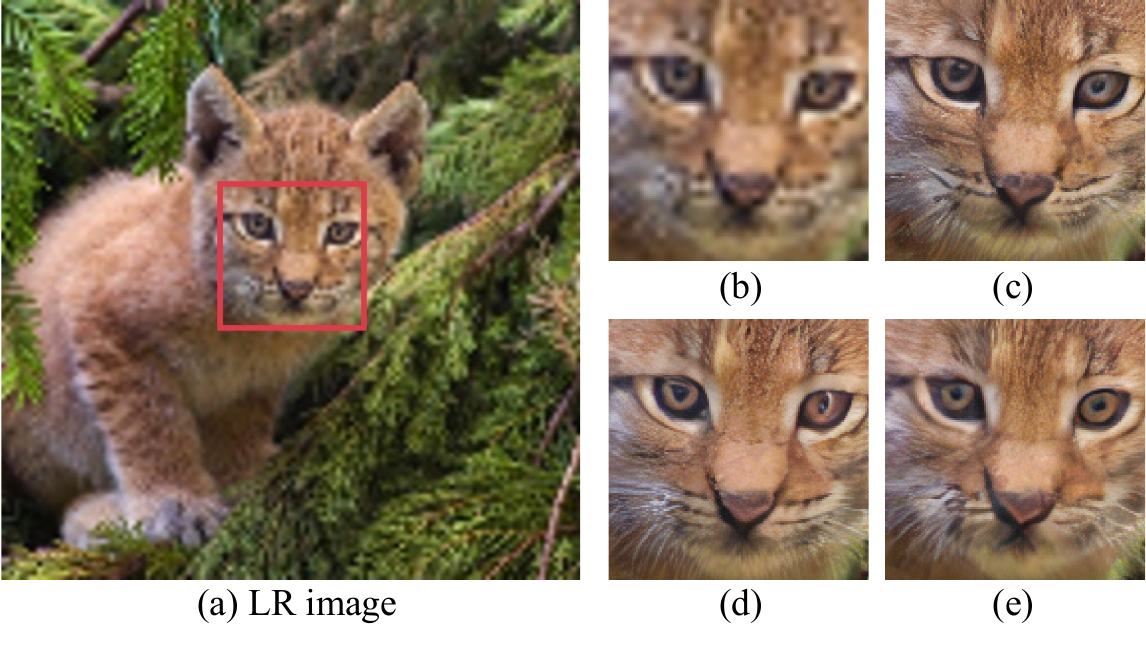}
    \caption{An illustration of the generative ability of the proposed method in only one step. Given the same LR image (Fig. (a) and (b)), by using different noise added to the input, HR images (Fig. (c)-(e)) with different details are generated, \eg, eyes of different shapes and colors. Best zoom in for details.}
    \label{fig:randomness}
    \vspace{-0.3cm}
\end{figure}

To address these \bihan{challenges}, we \bihan{introduce} a novel approach that can generate high-resolution (HR) images in only one sampling step, without \bihan{compromising} the diversity and perceptual quality of the diffusion model, as shown in Fig.~\ref{fig:intro} and Fig.~\ref{fig:randomness}. 
Specifically, we propose to directly learn a well-paired bi-directional deterministic mapping between the input random noise and the generated HR image from a teacher diffusion model. 
%
%
To accelerate the generation of well-matched training data, we first derive a deterministic sampling strategy 
\bihan{from the most recent state-of-the-art work \cite{yue2023resshift}, designed for accelerating diffusion-based SR, from its original stochastic formulation.}
%
\bihan{Additionally, we propose a novel consistency-preserving loss to leverage ground-truth images, further enhancing the perceptual quality of the generated HR images by minimizing the error between ground-truth (GT) images and those generated from the predicted initial state.}
Experimental results demonstrate that our method achieves comparable or even better performance \bihan{compared to} SOTA methods and the teacher diffusion model~\cite{yue2023resshift}, while greatly reducing the number of inference steps from $15$ to $1$, \bihan{resulting in} up to a $\times10$ speedup in inference.

Our main contributions are summarized as follows:
\begin{itemize}
    \item 
    We accelerate the diffusion-based SR model to a single inference step with comparable or even superior performance for the first time.
    Instead of shortening the Markov chain of the generation process, we propose a simple yet effective approach that directly distills a deterministic generation function into a student network.
    \item 
    To further fasten training, we derive a deterministic sampling strategy from the recent SOTA method \cite{yue2023resshift} on accelerating the SR task, enabling efficient generation of well-matched training pairs.
    \item 
    We propose a novel consistency-preserving loss that can utilize the ground-truth images during training, preventing the student model from only focusing on fitting the deterministic mapping of the teacher diffusion model, therefore leading to better performance.
    \item 
    Extensive experiments on both synthetic and real-world datasets \bihan{show} that our proposed method can achieve comparable or even superior performance \bihan{compared to} SOTA methods and the teacher diffusion model, while greatly reducing the number of inference steps from $15$ to $1$.
\end{itemize}

\vspace{-0.2cm}
\section{Related Work}
\subsection{Image Super-Resolution} 
With the rise of deep learning, deep learning-based techniques gradually become the mainstream of the SR task~\cite{dong2015image, wang2020deep}. One prevalent approach of early works is to train a regression model using paired training data~\cite{Alpher02, ahn2018image, Wang_Liu_Yang_Han_Huang_2015, Kim_Lee_Lee_2016}. While the expectation of the posterior distribution can be well modeled, they inevitably suffer from the over-smooth problem \cite{ledig2017photo, Sajjadi_Scholkopf_Hirsch_2017, Menon_Damian_Hu_Ravi_Rudin_2020}. 
To improve the perceptual quality of the generated HR images, generative-based SR models attract increasing attention, \eg, autoregressive-based models~\cite{Dahl_Norouzi_Shlens_2017, Menick_Kalchbrenner_2018, Oord_Kalchbrenner_Vinyals_Espeholt_Graves_Kavukcuoglu_2016, Parmar_Vaswani_Uszkoreit_Kaiser_Shazeer_Ku_Tran_2018}. While significant improvements are achieved, the computational cost of autoregressive models is usually large. 
Subsequently, normalizing flows~\cite{lugmayr2020srflow, wang2022low} are demonstrated to have good perceptual quality under an efficient inference process, while its network design is restricted by the requirements of the invertibility and ease of calculation. Besides, GAN-based methods also achieve great success in terms of perceptual quality~\cite{ledig2017photo, Sajjadi_Scholkopf_Hirsch_2017, Menon_Damian_Hu_Ravi_Rudin_2020, guo2022lar, karras2018progressive}.
However, the training of GAN-based methods is usually unstable. Recently, diffusion-based models have been widely investigated in SR~\cite{saharia2022image, rombach2022high, kawar2022denoising, chung2022come, Choi_Kim_Jeong_Gwon_Yoon_2021}. 
The diffusion-based SR methods can be roughly summarized into two categories, concatenating the LR image to the input of the denoiser~\cite{saharia2022image, rombach2022high}, and modifying the backward process of a pre-trained diffusion model~\cite{kawar2022denoising, chung2022come, Choi_Kim_Jeong_Gwon_Yoon_2021}. Although promising results are achieved, they rely on a large number of inference steps, which greatly hinders the application of diffusion-based models.

\subsection{Acceleration of Diffusion Models}
Recently, the acceleration of diffusion models has attracted more and more attention. Several algorithms are proposed for general diffusion models~\cite{song2020denoising, nichol2021improved, lu2022dpm, song2023consistency} and proved quite effective for image generations. One intuitive strategy among them is to distill the diffusion models to a student model. However, the huge training overhead to solve the ordinary differential equation (ODE) of the inference process makes this scheme less attractive on a large-scale dataset~\cite{luhman2021knowledge}. 
To alleviate the training overhead, progressive distillation strategies are usually adopted~\cite{salimans2021progressive, meng2023distillation}. Meanwhile, instead of simply simulating the behavior of a teacher diffusion model through distillation, better inference paths are explored in an iterative manner~\cite{liu2022flow, lipman2023flow}. While progressive distillation effectively decreases the training overhead, the error accumulates at the same time, leading to an obvious performance loss in SR. 
Most recently, targeting the image restoration task, some works reformulate the diffusion process by either using the knowledge of degradation process~\cite{wang2023exposurediffusion} or a pre-defined distribution of the initial state~\cite{yue2023resshift}, yielding a shortened Markov chain of the generation process and better performance than directly applying DDIM~\cite{song2020denoising} in low-level tasks. However, they either require a clear formulation of the degradation or still require a relatively large number of inference steps.

\begin{figure}
    \centering
    \begin{subfigure}{1\linewidth}
        \includegraphics[width=\linewidth]{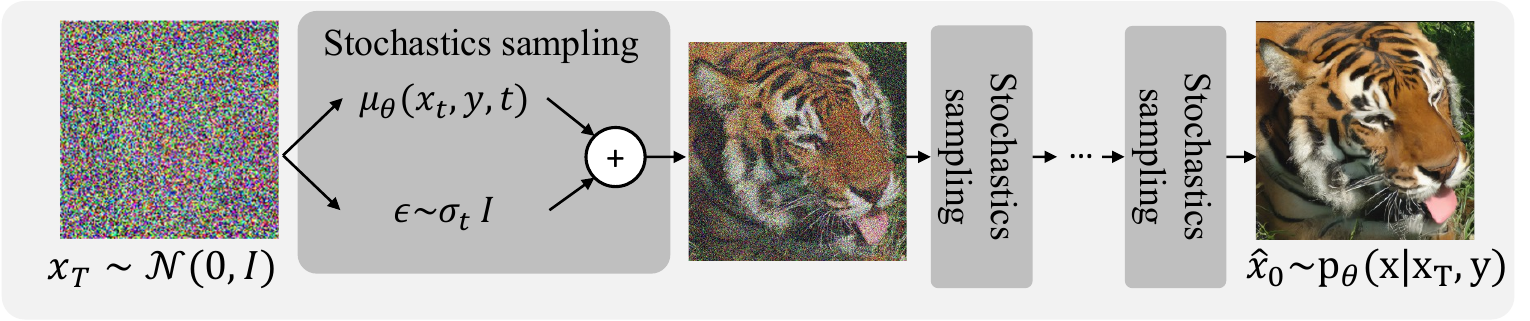}
        \caption{The inference of SR3~\cite{saharia2022image} starts from a pure noise, which requires a large number of inference steps (T=$100$ after using DDIM~\cite{song2020denoising}).}
    \end{subfigure}
    \begin{subfigure}{1\linewidth}
        \includegraphics[width=\linewidth]{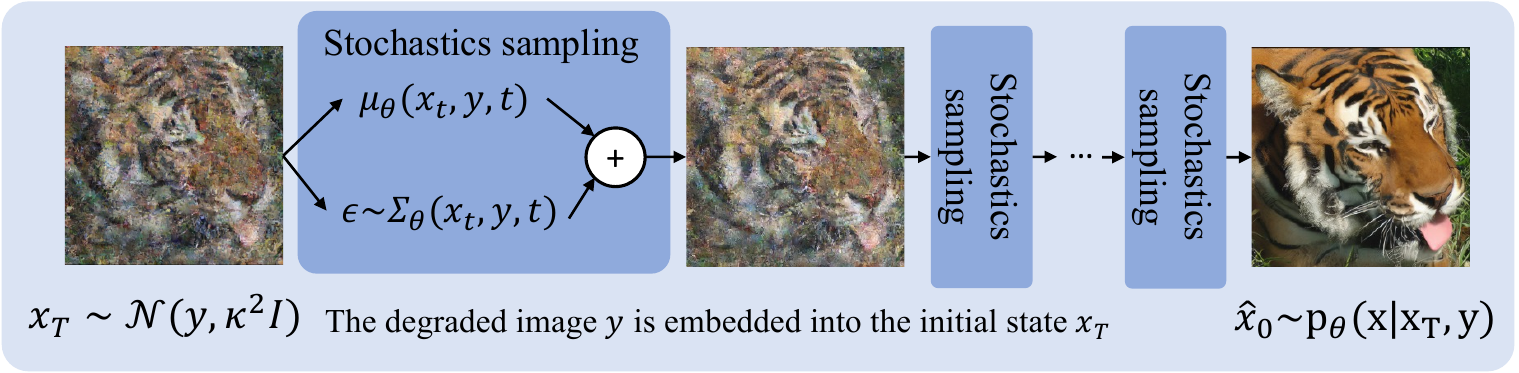}
        \caption{The recent SOTA method ResShift~\cite{yue2023resshift} shortens the Markov chain to speed up the inference process by incorporating the information of the LR image $y$ to the initial state $x_T$ (T=$15$).}
    \end{subfigure}
        \begin{subfigure}{1\linewidth}
        \includegraphics[width=\linewidth,trim=0 0 10 0,clip]
        {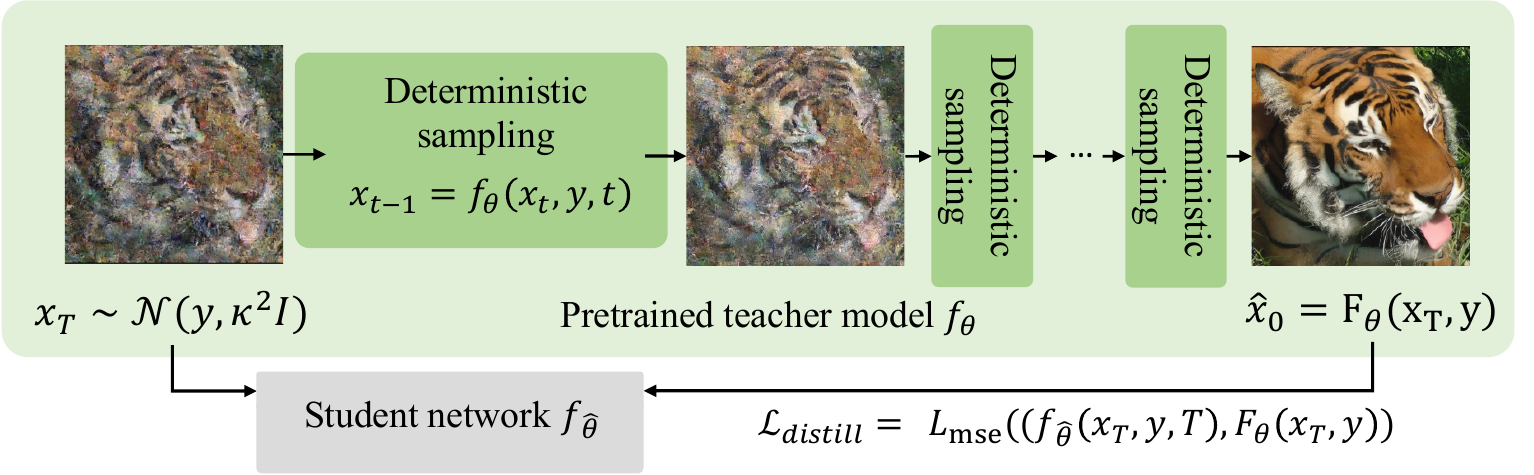}
        \caption{A simplified pipeline of the proposed method \textit{SinSR} (distill only). It directly learns the deterministic mapping between $x_T$ and $x_0$, therefore the inference process can be further compressed into only one step (T=$1$).}
    \end{subfigure}
    \vspace{-0.6cm}
    \caption{A comparison between the vanilla diffusion-based SR method~\cite{saharia2022image}, a most recent method for acceleration of the diffusion-based SR~\cite{yue2023resshift}, and the proposed one-step SR. Different from recent works that shorten the Markov chain to speed up the inference process~\cite{wang2023exposurediffusion, yue2023resshift}, the proposed method directly learns the deterministic generation process and the details can be found in Fig. \ref{fig:framework}.}
    \label{fig:compare_method}
    \vspace{-0.4cm}
\end{figure}

\begin{figure*}[t]
    \centering
    \vspace{-0.3cm}
    \includegraphics[width=1\linewidth]{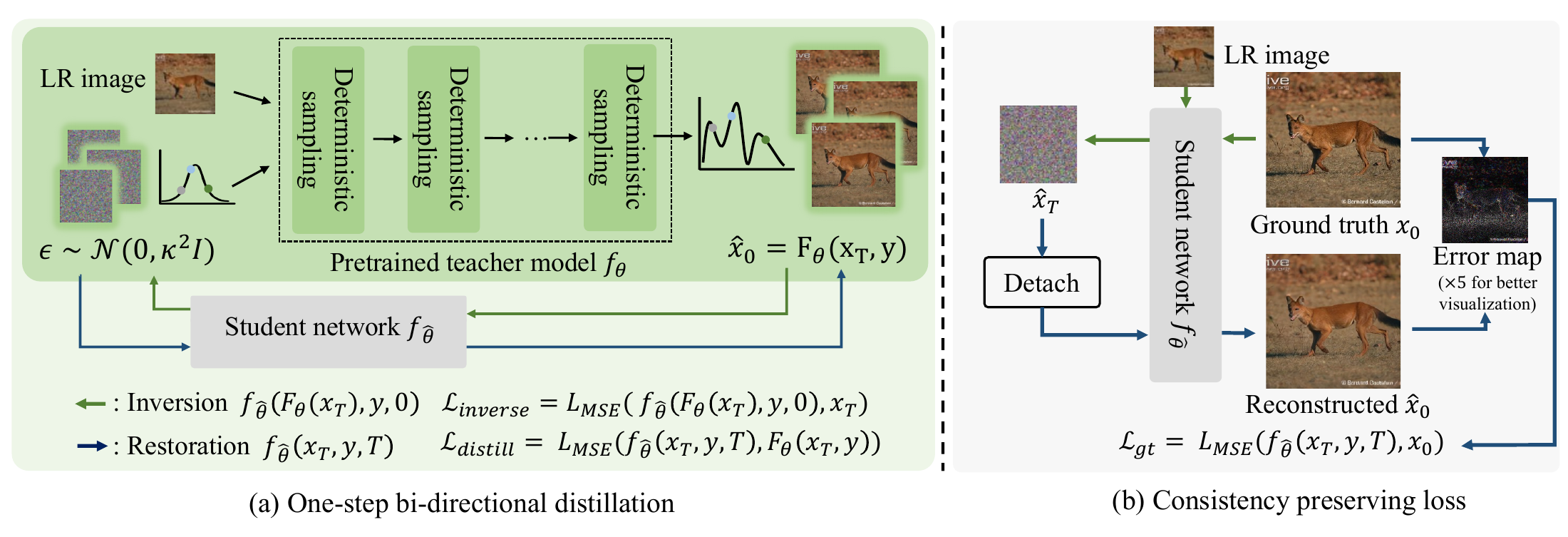}
    \vspace{-0.8cm}
    \caption{The overall framework of the proposed method. By minimizing $\mathcal{L}_{distill}$ and $\mathcal{L}_{inverse}$, the student network $f_{\hat{\theta}}$ learns the deterministic bi-directional mapping between $x_T$ and $\hat{x}_0$ obtained from a pre-trained teacher diffusion model in one step. Meanwhile, the proposed consistency preserving loss $\mathcal{L}_{gt}$ is optimized during training to utilize the information from the GT images to pursue better perceptual quality instead of simply fitting the deterministic mappings from the teacher model. Specifically, the GT image is first converted to its latent code $\hat{x}_T=f_{\hat{\theta}}(x_{0}, y, 0)$, and then converted back to calculate its reconstruction loss $L_{MSE}(f_{\hat{\theta}}(\hat{x}_T, y, T), x_0)$.}
    \label{fig:framework}
    \vspace{-0.25cm}
\end{figure*}

\section{Motivation}
\noindent\textbf{Preliminary.}
Given \bihan{an} LR image $y$ and its corresponding HR image $x_0$, existing diffusion-based SR methods aim to model the conditional distribution $q(x_0|y)$ through a Markov chain where a forward process is usually defined as $q(x_t|x_{t-1})=\mathcal{N}(x_t; \sqrt{1-\beta_t} x_{t-1}, \beta_t I)$ with an initial state $x_T\sim \mathcal{N}(0, I)$.
The role of the diffusion model can be regarded as transferring the input domain (standard Gaussian noise) to the HR image domain conditioned on the LR image. Since the matching relationship between $x_T$ and $x_0$ is unknown, usually a diffusion model~\cite{ddpm, saharia2022image, liu2022flow} through an iterative manner is required to learn/infer from an unknown mapping between $x_T$ and $x_0$. 
\bihan{Our method is grounded in the idea that having an SR model that effectively captures the conditional distribution $q(x_0|y)$ and establishes a deterministic mapping between $x_T$ and $\hat{x}0$ given an LR image $y$, we can streamline the inference process to a single step by employing another network, denoted as $f_{\hat{\theta}}$, to learn the correspondence between $\hat{x}_0$ and $x_T$, as illustrated in Fig. \ref{fig:compare_method}.}

\noindent\textbf{Distillation for diffusion SR models: less is more.}
While the \bihan{concept} of distilling the mapping between $x_T$ and $\hat{x}_0$ to a student network has been previously explored~\cite{liu2022flow}, 
\bihan{its application to SR introduces several challenges:}
\begin{itemize}
    \item The training overhead \bihan{becomes substantial} for one-step distillation due to a large number of inference steps of previous models, \eg, LDM~\cite{rombach2022high} still need $100$ steps after using DDIM~\cite{song2020denoising} for inference to generate high-quality pairs $(\hat{x}_0, x_T, y)$ as the training data of the student model.
    \item The performance degradation 
    \bihan{is attributed to the introduction of a more intricate distillation strategy involving iteration.}
    For example, to reduce the training overhead, an iterative distillation strategy~\cite{salimans2021progressive} is adopted which gradually decreases the number of inference steps during training. However, \bihan{despite achieving} satisfactory results in generation tasks, the \bihan{cumulative} error significantly impacts the fidelity of the SR results, as SR tasks are relatively more sensitive to image quality.
\end{itemize}

To address the aforementioned two challenges, we propose to distill the diffusion SR process into a single step in a simple but effective way based on the following observations. More details of the observations can be seen in Sec. \ref{sec:ana}

\begin{itemize}
    \item We demonstrate that the most recent SOTA method for accelerating the diffusion-based SR~\cite{yue2023resshift}, which achieves comparable performance in $15$ steps as LDM~\cite{rombach2022high} in $100$ DDIM steps, has a deterministic mapping between $x_T$ and $x_0$. Besides, the greatly reduced number of inference steps and the existence of the deterministic mapping make the training of a single-step distillation possible as shown in Fig.~\ref{fig:ablation_ode} and Table \ref{tab:sampling}.  
    \item Learning the mapping between $x_T$ and $\hat{x}_0$ is found to be easier than denoising $x_t$ under different noise levels as shown in Table~\ref{tab:ablation_small}. Therefore, it is feasible to directly learn the mapping between $x_T$ and $\hat{x}_0$ so that the accumulated error by the iterative distillation can be avoided.
    \item Due to the accumulated error, a more sophisticated distillation strategy (iterative-based) does not contribute to the improvement in our setting as shown in Table~\ref{tab:compare_rectified}.
\end{itemize}


The organization of the following sections is as follows: we first demonstrate that ResShift~\cite{yue2023resshift}, in which the inference process is originally stochastic, can be converted to a deterministic model without retraining in Sec~\ref{sec:deter}, and then the proposed consistency preserving distillation in Sec~\ref{sec:distill}.

\section{Methodology}
\subsection{Deterministic Sampling}
\label{sec:deter}
A core difference between ResShift~\cite{yue2023resshift} and LDM~\cite{rombach2022high} is the formulation of the initial state $x_T$. Specifically, in ResShift~\cite{yue2023resshift}, the information from the LR image $y$ is integrated into the diffusion step $x_t$ as follows
\footnote{For ease of presentation, the LR image is $y$ is pre-upsampled to the same spatial resolution with the HR image $x$. Besides, similar to \cite{yue2023resshift, rombach2022high}, the diffusion is conducted in the latent space.}
\begin{equation}
q(x_t|x_0,y)=\mathcal{N}(x_t; x_0+\eta_t (y-x_0), \kappa^2\eta_t \mathbf{I}),    
\end{equation}
where $\eta_t$ is a serial of hyper-parameters that monotonically increases with the timestep $t$ and obeys $\eta_T \to 1$ and $\eta_0 \to 0$. As such, the inverse of the diffusion process starts from an initial state with rich information from the LR image $y$ as follows $x_T=y+\kappa \sqrt{\eta_T} \epsilon$ where $\epsilon\sim \mathcal{N}(\mathbf{0}, \mathbf{I})$.
To generate a HR image $x$ from a given image $y$, the original inverse process of~\cite{yue2023resshift} is as follows
\begin{equation}
p_\theta(x_{t-1}|x_t, y)=\mathcal{N}(x_{t-1}|\mu_{\theta}(x_t, y, t), \kappa^2\frac{\eta_{t-1}}{\eta_t}\alpha_t \mathbf{I}) ,
\label{eq:sample}
\end{equation}
where $\mu_{\theta}(x_t, y, t)$ is reparameterized by a deep network. As shown in Eq.~\ref{eq:sample}, given an initial state $x_T=y+\kappa \sqrt{\eta_T} \epsilon$, the generated image is stochastic due to the existence of the random noise during the sampling from $p_\theta(x_{t-1}|x_t, y)$. Inspired by DDIM sampling~\cite{song2020denoising}, we find that a non-Markovian reverse process $q(x_{t-1}|x_t, x_0, y)$ exists which keeps the marginal distribution $q(x_t|x_0, y)$ unchanged so that it can be directly adopted to a pre-trained model. The reformulated deterministic reverse process is as follows
\begin{equation}
    q(x_{t-1}|x_t, x_0, y)= \delta({k_t x_0 + m_t x_t + j_t y}) ,
\end{equation}
where $\delta$ is the unit impulse, and $k_t, m_t, j_t$ are as follows
\begin{equation}
    \left\{\begin{matrix}
m_t = \sqrt{\frac{\eta_{t-1}}{\eta_t}} \\ 
j_t = \eta_{t-1} - \sqrt{\eta_{t-1}\eta_{t}} \\
k_t = 1 - \eta_{t-1} + \sqrt{\eta_{t-1}\eta_{t}} - \sqrt{\frac{\eta_{t-1}}{\eta_t}}.
\end{matrix}\right.
\end{equation}
The details of the derivation can be found in the supplementary material. As a consequence, for inference, the reverse process conditioned on $y$ is reformulated as follows 
\begin{equation}
\begin{split}
    x_{t-1} &= k_t \hat{x}_0 + m_t x_t + j_t y \\
    &= k_t f_\theta(x_t, y, t) + m_t x_t+ j_t y ,
\end{split}
\label{eq:deter_sample}
\end{equation}
where $f_\theta(x_t, y, t)$ is the predicted HR image from a pre-trained ResShift~\cite{yue2023resshift} model. By sampling from the reformulated process in Eq. \ref{eq:deter_sample}, a deterministic mapping between $x_T$ (or $\epsilon$) and $\hat{x}_0$ can be obtained and is denoted as $F_\theta(x_T, y)$.

\subsection{Consistency Preserving Distillation}
\label{sec:distill}
\noindent\textbf{Vanilla distillation.}
We propose utilizing a student network  $f_{\hat{\theta}}$ to learn the deterministic mapping $F_{\theta}$ between the random initialized state $x_T$ and its deterministic output $F_{\theta}(x_T, y)$ from a teacher diffusion model. The vanilla distillation loss is defined as follows
\begin{equation}
    \mathcal{L}_{distill} =  L_{MSE}(f_{\hat{\theta}}(x_T, y, T), F_{\theta}(x_T, y)),
\label{eq:matching}
\end{equation}
where $f_{\hat{\theta}}(x_T, y, T)$ is the student network that directly predicts the HR image in only one step, and $F_{\theta}$ represents the proposed deterministic inference process of ResShift~\cite{yue2023resshift} in Sec.~\ref{sec:deter} through an iterative manner using a pre-trained network parameterized by $\theta$. We observe that the student model trained solely with the distillation loss in Eq.~\ref{eq:matching} already achieves promising results in just one inference step, as indicated by ``(distill only)'' in the result tables.

\noindent\textbf{Regularization by the ground-truth image.}
A limitation of the aforementioned vanilla distillation strategy is that the GT image is not utilized during training, thereby restricting the upper performance bound of the student model. To further enhance the student's performance, we propose a novel strategy that incorporates a learned inversion of the HR image to provide additional regularization from the ground-truth images. In addition to the vanilla distillation loss, the student network concurrently learns the inverse mapping during training by minimizing the following loss,
\begin{equation}
    \mathcal{L}_{inverse} =  L_{MSE}(f_{\hat{\theta}}(F_{\theta}(x_T, y), y, 0), x_T),
\label{eq:reverse}
\end{equation}
where the last parameter of $f_{\hat{\theta}}$ is set from $T$ in Eq. \ref{eq:matching} to $0$, indicating that the model is predicting the inversion instead of the $\hat{x}_0$. Then the GT image $x_0$ can be employed to regularize the output SR image given its predicted inversion $\hat{x}_T$ as follows
\begin{equation}
\begin{split}
    \hat{x}_T &= detach(f_{\hat{\theta}}(x_0, y, 0)) \\
    \mathcal{L}_{gt} &=  L_{MSE}(f_{\hat{\theta}}(\hat{x}_T, y, T), x_0),
\end{split}
\label{eq:gt_loss}
\end{equation}
where $\mathcal{L}_{gt}$ is the proposed consistency preserving loss. By reusing $f_{\hat{\theta}}$ to learn both $f_{\hat{\theta}}(\cdot, \cdot, T)$ and $f_{\hat{\theta}}(\cdot, \cdot, 0)$ simultaneously, we can initialize the parameter $\hat{\theta}$ of the student model from the teacher one $\theta$ to speed up the training.

\noindent\textbf{The overall training objective.}
The student network is trained to minimize the aforementioned three losses at the same time as follows
\begin{equation}
    \hat{\theta} = \arg\min_{\hat{\theta}} \mathbb{E}_{y, x_0, x_T} [ \mathcal{L}_{distill} + \mathcal{L}_{reverse} + \mathcal{L}_{gt}],  
\end{equation}
where the losses are defined in Eq. \ref{eq:matching}, \ref{eq:reverse}, and \ref{eq:gt_loss} respectively. We assign equal weight to each loss term, and ablation studies are in the supplementary material. The overall of the proposed method is summarized in Algorithm \ref{alg:training} and Fig. \ref{fig:framework}.

\begin{algorithm}[t]
    \caption{Training}
    \scalebox{1}{
    \begin{minipage}{1\linewidth}
    \begin{algorithmic}[1]
    \REQUIRE Pre-trained teacher diffusion model $f_\theta$
    \REQUIRE Paired training set $(X, Y)$
    
    \STATE Init $f_{\hat{\theta}}$ from the pre-trained model, \ie, $\hat{\theta} \leftarrow  \theta$.
    \WHILE{not converged}
        \STATE sample $x_0, y \sim (X, Y)$
        \STATE sample $\epsilon \sim \mathcal{N}(\mathbf{0}, \kappa^2 {\eta_T} \mathbf{I})$
        \STATE $x_T = y+\epsilon$
        \FOR {$t = T, T-1, ..., 1$} 
        \IF{$t=1$}
            \STATE $\hat{x}_{0}=f_\theta(x_1, y, 1)$
        \ELSE
            \STATE $x_{t-1}= k_t f_\theta (x_t, y, t) + m_t x_t+ j_t y$ 
        \ENDIF
        \ENDFOR        
        \STATE $\mathcal{L}_{distill} =  L_{MSE}(f_{\hat{\theta}}(x_T, y, T), \hat{x}_0)$
        \STATE $\mathcal{L}_{inverse} =  L_{MSE}(f_{\hat{\theta}}(\hat{x}_0, y, 0), x_T)$
        \STATE $\hat{x}_T = f_{\hat{\theta}}(x_0, y, 0)$, 
        \STATE $\mathcal{L}_{gt} = L_{MSE}(f_{\hat{\theta}}({detach}(\hat{x}_T), y, T), x_0)$
        \STATE $\mathcal{L} = \mathcal{L}_{distill}+\mathcal{L}_{inverse}+\mathcal{L}_{gt}$
        \STATE Perform a gradient descent step on $\nabla_{\hat{\theta}} \mathcal{L} $ 
    \ENDWHILE
    \RETURN The student model $f_{\hat{\theta}}$.
    \end{algorithmic}
    \end{minipage}
    }
    \label{alg:training}
\end{algorithm}
\section{Experiment}
\subsection{Experimental setup}

\noindent\textbf{Training Details.} For a fair comparison, we follow the same experimental setup and backbone design as that in \cite{yue2023resshift}. Specifically, the main difference is that we finetuned the model for 30K iterations instead of training from scratch for 500K in \cite{yue2023resshift}. We find that the student model can converge quickly so that even if for each iteration we need extra time to solve the ODE to get paired training data, the overall training time is still much shorter than retraining a model from scratch following \cite{yue2023resshift}. We train the models on the training set of ImageNet~\cite{deng2009imagenet} following the same pipeline with ResShift~\cite{yue2023resshift} where the degradation model is adopted from RealESRGAN~\cite{wang2021real}. 

\noindent\textbf{Compared methods.} We compare our method with several representative SR models, including RealSR-JPEG~\cite{ji2020real}, ESRGAN~\cite{wang2018esrgan}, BSRGAN~\cite{zhang2021designing}, SwinIR~\cite{liang2021swinir}, RealESRGAN~\cite{wang2021real}, DASR~\cite{liang2022efficient}, LDM~\cite{rombach2022high}, and ResShift~\cite{yue2023resshift}. For a comprehensive comparison, we further evaluate the performance of diffusion-based models LDM~\cite{rombach2022high} and ResShift~\cite{yue2023resshift} with a reduced number of sampling steps. Besides, we compare the proposed method with RectifiedFlow~\cite{liu2022flow}, a SOTA method that can compress the generation process into a single step, in Table~\ref{tab:compare_rectified}.

\noindent\textbf{Metrics.}
For the evaluation of the proposed method on the synthetic testing dataset with reference images, we utilize PSNR, SSIM, and LPIPS~\cite{zhang2018unreasonable} to measure the fidelity performance. Besides, two recent SOTA non-reference metrics are used to justify the realism of all the images, \ie, CLIPIQA~\cite{wang2023exploring} which leverages a CLIP model~\cite{radford2021learning} pre-trained on a large-scale dataset (Laion400M~\cite{schuhmann2021laion}) and MUSIQ~\cite{ke2021musiq}.

\subsection{Experimental Results}
\begin{figure*}
\vspace{-0.3cm}
\centering
\includegraphics[width=\linewidth]{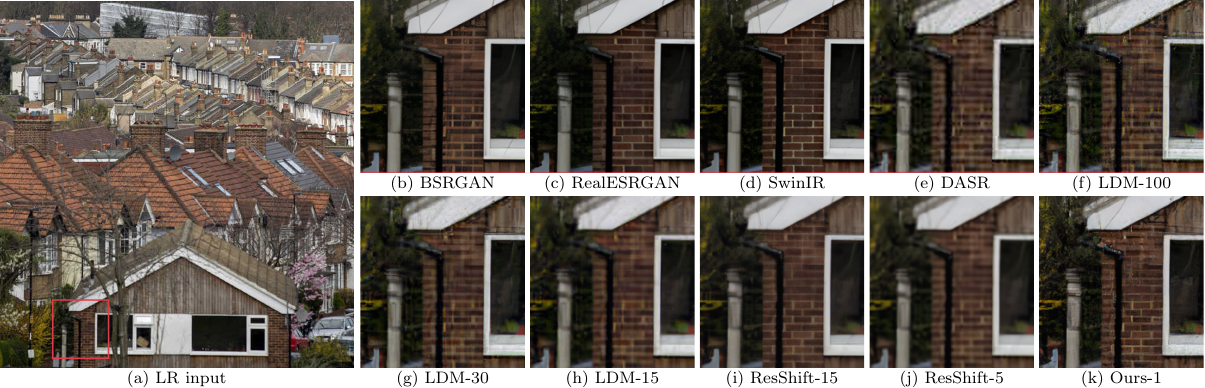}
\vspace{0.1cm}
\includegraphics[width=\linewidth]{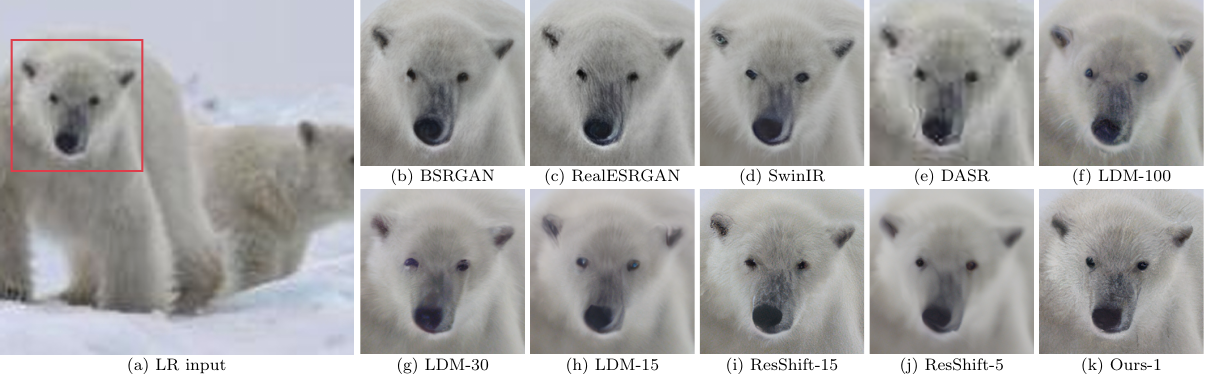}
\vspace{0.1cm}
\includegraphics[width=\linewidth]{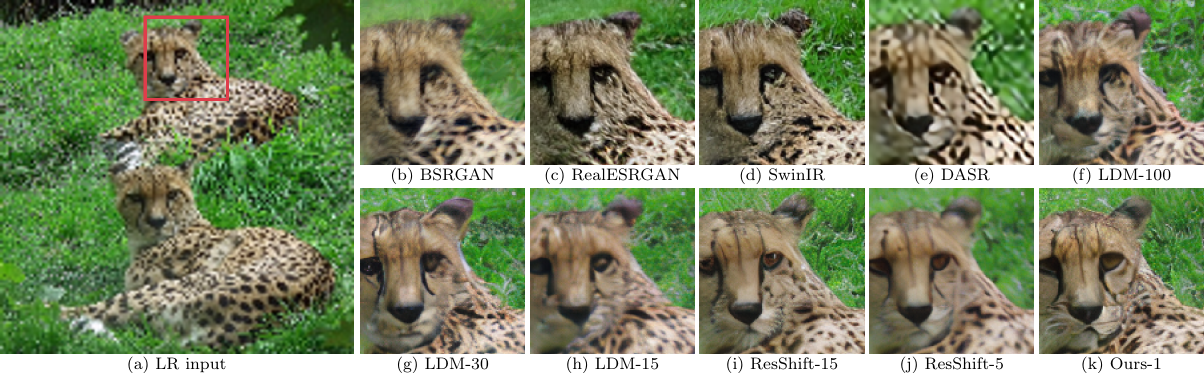}
\vspace{-0.7cm}
\caption{Visual comparison on real-world samples. Please zoom in for more details.}
\label{fig:main_res}
\vspace{-0.3cm}
\end{figure*}

\begin{table*}[t]
    \centering
    \small
    \vspace{-0.4cm}
    \begin{tabular}{@{}C{4.0cm}@{}|
                    @{}C{2.4cm}@{} @{}C{2.5cm}@{}| 
                    @{}C{2.5cm}@{} @{}C{2.5cm}@{} }
        \Xhline{0.8pt}
        \multirow{3}*{Methods} & \multicolumn{4}{c}{Datasets} \\
        \Xcline{2-5}{0.4pt}
            & \multicolumn{2}{c|}{\textit{RealSR}}  & \multicolumn{2}{c}{\textit{RealSet65}} \\
            \Xcline{2-5}{0.4pt}
            & CLIPIQA$\uparrow$ & MUSIQ$\uparrow$   
            & CLIPIQA$\uparrow$ & MUSIQ$\uparrow$  \\
            \Xhline{0.4pt}
            ESRGAN~\cite{wang2018esrgan}     & 0.2362 & 29.048   & 0.3739 & 42.369  \\
            RealSR-JPEG~\cite{ji2020real}    & 0.3615 & 36.076  & 0.5282 & 50.539  \\
            BSRGAN~\cite{zhang2021designing} & {0.5439} & \textbf{63.586} & 0.6163 &\textbf{65.582} \\
            SwinIR~\cite{liang2021swinir}    & 0.4654 & 59.636  & 0.5782 & \underline{63.822} \\
            RealESRGAN~\cite{wang2021real}   & 0.4898 & 59.678  & 0.5995 & 63.220  \\
            DASR~\cite{liang2022efficient}   & 0.3629 & 45.825  & 0.4965 & 55.708   \\
            LDM-15~\cite{rombach2022high}    & 0.3836 & 49.317 & 0.4274 & 47.488  \\
            \hline
            ResShift-15~\cite{yue2023resshift}       & {0.5958} & {59.873} & {0.6537} & 61.330  \\
            \textit{SinSR-1} (distill only) & \underline{0.6119} & 57.118 & \underline{0.6822} & 61.267 \\
            \textit{SinSR-1} & \textbf{0.6887} & \underline{61.582} & \textbf{0.7150} & 62.169 \\
       \Xhline{0.4pt}
    \end{tabular} 
    \vspace{-0.25cm}
    \caption{Quantitative results of models on two real-world datasets. The best and second best results are highlighted in \textbf{bold} and \underline{underline}.}
    \label{tab:real_testing}
\end{table*}

\begin{table*}[t]
    \centering
    \small
    \vspace{-2mm}
    \begin{threeparttable}
    \begin{tabular}{@{}C{3.8cm}@{}|
                    @{}C{1.9cm}@{} @{}C{2.0cm}@{} @{}C{2.0cm}@{} @{}C{2.1cm}@{} @{}C{2.1cm}@{}}
        \Xhline{0.8pt}
        \multirow{2}*{Methods} & \multicolumn{5}{c}{Metrics} \\
        \Xcline{2-6}{0.4pt}
            & PSNR$\uparrow$ & SSIM$\uparrow$ & LPIPS$\downarrow$ & CLIPIQA$\uparrow$ & MUSIQ$\uparrow$ \\
            \Xhline{0.4pt}
            ESRGAN~\cite{wang2018esrgan}     & 20.67 & 0.448 & 0.485 & 0.451 & 43.615  \\
            RealSR-JPEG~\cite{ji2020real}    & 23.11 & 0.591 & 0.326 & 0.537 & 46.981  \\
            BSRGAN~\cite{zhang2021designing} & 24.42 & 0.659 & 0.259 & {0.581} & \textbf{54.697}  \\
            SwinIR~\cite{liang2021swinir}    & 23.99 & 0.667 & {0.238} & 0.564 & {53.790} \\
            RealESRGAN~\cite{wang2021real}   & 24.04 & 0.665 & 0.254 & 0.523 & 52.538 \\
            DASR~\cite{liang2022efficient}   & 24.75 & \textbf{0.675} & 0.250 & 0.536 & 48.337  \\
            LDM-30~\cite{rombach2022high}  & 24.49 & 0.651 & 0.248 & 0.572 & 50.895  \\
            LDM-15~\cite{rombach2022high}  & \underline{24.89} & 0.670 & 0.269 & 0.512 & 46.419  \\
            \hline
            ResShift-15~\cite{yue2023resshift} & \textbf{24.90} & \underline{0.673} & {0.228} & {0.603} & \underline{53.897}  \\
            \textit{SinSR-1} (distill only)  & 24.69 & 0.664 & \underline{0.222} & \underline{0.607} & 53.316 \\
            \textit{SinSR-1} & 24.56 & 0.657 & \textbf{0.221} & \textbf{0.611} & 53.357 \\
       \Xhline{0.8pt}
    \end{tabular} 
    \end{threeparttable}
    \vspace{-0.25cm}
    \caption{Quantitative results of models on \textit{ImageNet-Test}. The best and second best results are highlighted in \textbf{bold} and \underline{underline}.}
    \label{tab:imagenet_testing}
    \vspace{-0.3cm}
\end{table*}

\begin{table*}[t]
    \centering
    \small
    \vspace{-0.3cm}
    \scalebox{0.90}{
    \begin{tabular}{@{}C{4cm}@{}|
                    @{}C{1.8cm}@{} @{}C{1.8cm}@{} @{}C{1.8cm}@{} | @{}C{1.8cm}@{} @{}C{1.8cm}@{}
                    @{}C{1.8cm}@{} @{}C{1.8cm}@{}|@{}C{1.8cm}@{}}
         \Xhline{0.8pt}
         \multirow{2}*{Metrics} & \multicolumn{8}{c}{Methods} \\
         \Xcline{2-9}{0.4pt}
                           & LDM-15 & LDM-30 & LDM-100  & ResShift-1 & ResShift-5 & ResShift-10 & ResShift-15 & \textit{SinSR-1} \\
        \Xhline{0.4pt}
         LPIPS$\downarrow$ & 0.269  & 0.248  & 0.244   & 0.383  & 0.345  &  0.274  & 0.228 &  \textbf{0.221}   \\
         CLIPIQA$\uparrow$ & 0.512  & 0.572  & \textbf{0.620}   &  0.340 & 0.417  &  0.512  & 0.603 &  \underline{0.611}    \\
         Runtime (bs=64) & 0.046s  & 0.080s  & 0.249s   & \textbf{0.012s} & 0.021s  &  0.033s & 0.047s &  \textbf{0.012s}  \\
         Runtime (bs=1) & 0.408s & 1.192s & 3.902s & \textbf{0.058s} & 0.218s & 0.425s & 0.633s & \textbf{0.058s} \\
         \Xhline{0.4pt}
         \# Parameters (M) & \multicolumn{3}{c|}{113.60}   & \multicolumn{4}{c|}{118.59} & 118.59   \\
        \Xhline{0.8pt}
    \end{tabular}}
    \vspace{-0.2cm}
    \caption{Efficiency and performance comparisons with SOTA methods on \textit{ImageNet-Test}. ``-N” represents the number of sampling steps the model used. The running time per image is tested on a Tesla A100 GPU on the x4 (64$\rightarrow$ 256) task averaged over the batch size (bs).}
    \label{tab:runtime_comparison}
    \vspace{-0.3cm}
\end{table*}

\noindent\textbf{Evaluation on real-world datasets.} RealSR~\cite{cai2019toward} and RealSet65~\cite{yue2023resshift} are adopted to evaluate the generalization ability of the model on unseen real-world data. Specifically, in RealSR~\cite{cai2019toward}, there are 100 real images captured by two different cameras in different scenarios. Besides, RealSet65~\cite{yue2023resshift} includes 65 LR images in total, collected from widely used datasets and the internet. 
The results on these two datasets are reported in Table~\ref{tab:real_testing}. As shown in the table, the proposed method with only one inference step can outperform the teacher model that we used by a large margin. Besides, for the latest metric CLIPIQA, the proposed method archives the best performance among all the competitors. Some visual comparisons are shown in Fig. \ref{fig:main_res}, in which the proposed method achieves promising results using only one step.

\noindent\textbf{Evaluation on synthetic datasets.} We further evaluate the performance of different methods on the synthetic dataset \textit{ImageNet-Test} following the setting in \cite{yue2023resshift}. Specifically, $3000$ high-resolution images are first randomly selected from the validation set of ImageNet~\cite{deng2009imagenet}. The corresponding LR images are obtained by using the provided script in \cite{yue2023resshift}. 
As shown in Table~\ref{tab:imagenet_testing}, while reducing the inference step from $15$ to only $1$ slightly decreases PSNR and SSIM, the proposed method achieves the best perceptual quality measured by LPIPS, a more recent full-reference image quality assessment (IQA) metric than SSIM. Besides, the proposed method also achieves the best performance among all the methods measured on the most recent SOTA metric CLIPIQA~\cite{wang2023exploring}, demonstrating that the proposed $1$-step model is on par with or even slightly better than the teacher model with 15 inference steps in terms of perceptual qualities.

\noindent\textbf{Evaluation of the efficiency.} 
We assess the computational efficiency of the proposed method in comparison to SOTA approaches. As shown in Table~\ref{tab:runtime_comparison}, the proposed method demonstrates superior performance with only one inference step, outperforming ResShift~\cite{yue2023resshift}—the adopted teacher model, which had already significantly reduced the inference time compared to LDM~\cite{rombach2022high}. It is worth noting that all methods presented in Table~\ref{tab:runtime_comparison} run in latent space, and the computational cost of VQ-VAE is counted.

\subsection{Analysis}
\label{sec:ana}
\begin{figure}[t]
    \centering
    \begin{subfigure}{0.32\linewidth}
    \includegraphics[width=\linewidth]{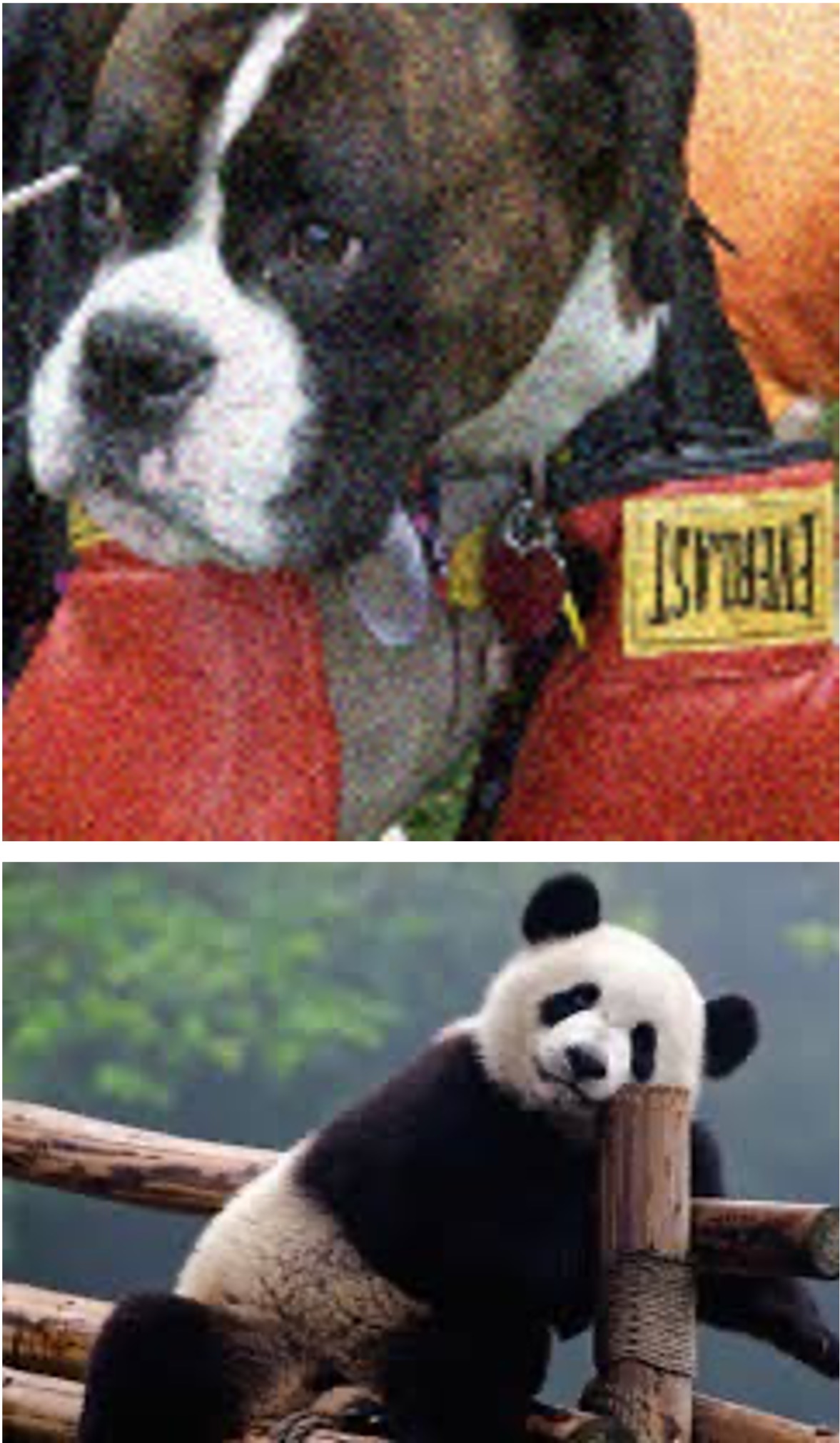}
    \caption{input}
    \end{subfigure}
    \begin{subfigure}{0.32\linewidth}
    \includegraphics[width=\linewidth]{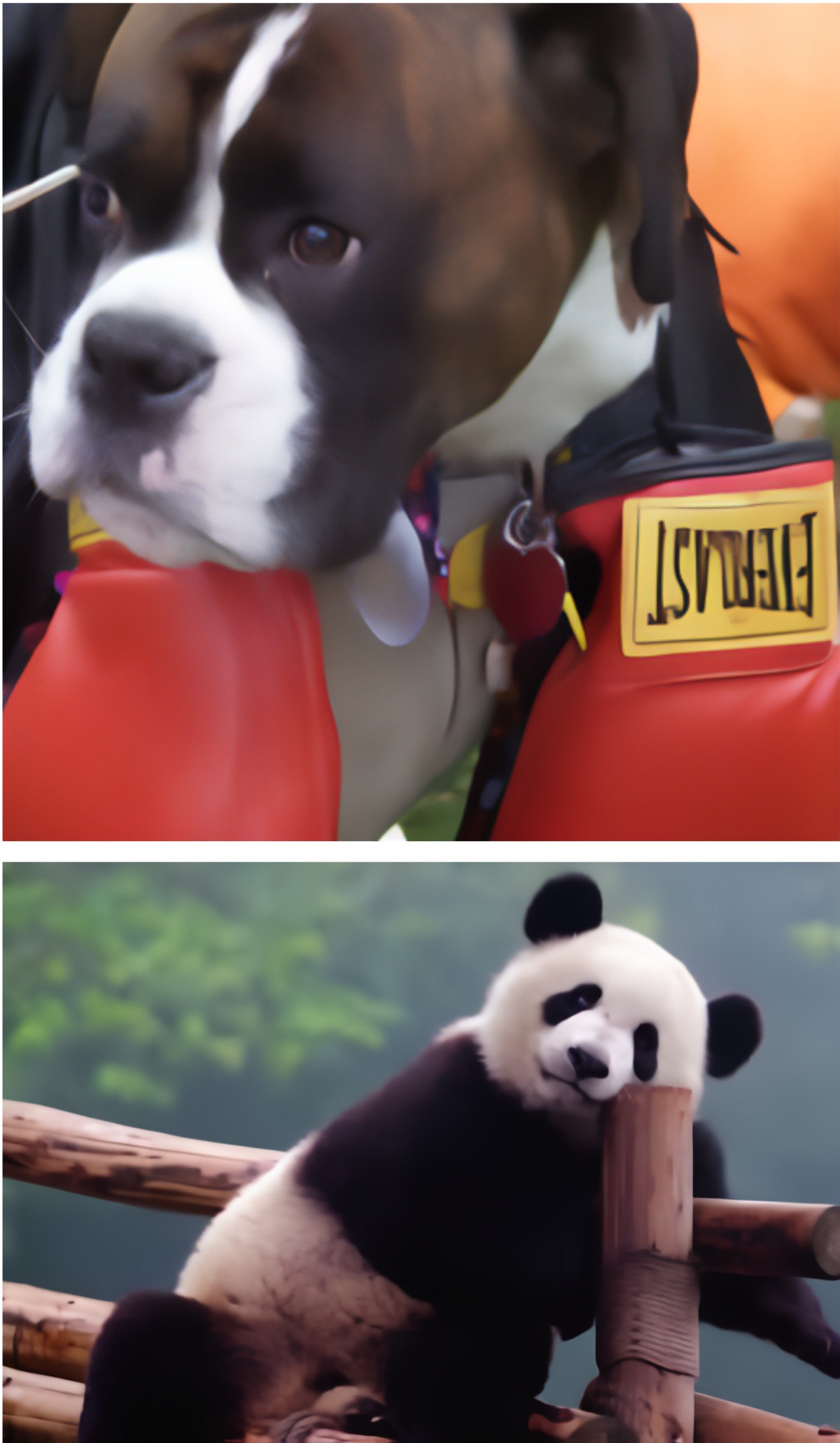}
    \caption{w/ sampling in~\cite{yue2023resshift}}
    \end{subfigure}
    \begin{subfigure}{0.32\linewidth}
    \includegraphics[width=\linewidth]{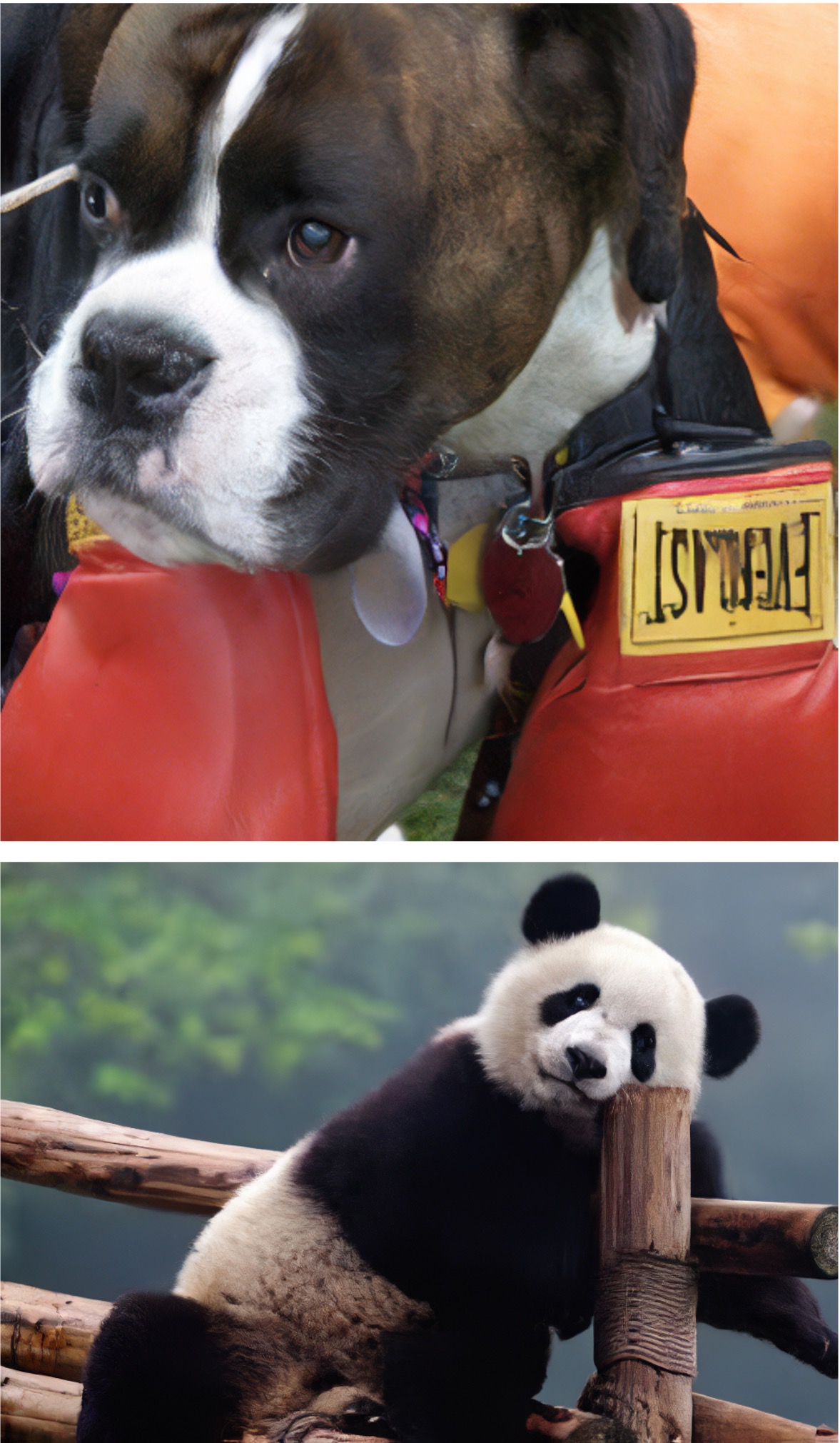}
    \caption{Ours (deterministic)}
    \end{subfigure}
    \vspace{-0.3cm}
    \caption{A comparison between the model trained with the default stochastic sampling process in ResShift~\cite{yue2023resshift} and the proposed deterministic sampling in Eq.~\ref{eq:deter_sample}. Best zoom in for more details.}
    \label{fig:ablation_ode}
    \vspace{-0.55cm}
\end{figure}

\noindent\textbf{How important is the deterministic sampling?} 
We evaluate the performance of the model trained on generated paired samples from the proposed deterministic sampling and the default stochastic sampling strategy $(x_T, \hat{F}_\theta(x_T, y))$ in~\cite{yue2023resshift}. 
Due to the randomness of the generated samples $x\sim \hat{F}_\theta(x_T, y)$, given a random noise $\epsilon$, the prediction is an expectation of its conditional distribution. The comparison in Fig. \ref{fig:ablation_ode} further verifies that the results trained w/o deterministic teacher model exhibit blurred details. 
Besides, as shown in Table~\ref{tab:sampling}, there is a significant performance degradation when we replace the proposed deterministic sampling with the default one in \cite{yue2023resshift}, demonstrating the effectiveness and necessity of involving the proposed deterministic sampling. 

\begin{table}[t]
    \centering
    \scalebox{0.9}{
    \begin{tabular}{ccc}
    \toprule
    Methods & CLIPIQA~$\uparrow$& MUSIQ~$\uparrow$ \\
    \midrule
    w/ default sampling in~\cite{yue2023resshift} & 0.4166 & 51.53 \\
    \textit{SinSR} (distill only) & \textbf{0.6822} &  \textbf{61.27}\\
    \bottomrule
    \end{tabular}}
    \vspace{-0.2cm}
    \caption{A comparison between the model trained with the default stochastic sampling process in ResShift~\cite{yue2023resshift} and the proposed deterministic sampling in Eq.~\ref{eq:deter_sample} using only distillation loss. We evaluate their performance on the RealSet65 testing set.}
    \label{tab:sampling}
    \vspace{-0.4cm}
\end{table}

\noindent\textbf{Why does a single-step distillation work?}
%
Previous studies suggest that directly learning the mapping between $x_T$ and $x_0$ is typically challenging due to the non-causal properties of the generation process~\cite{lipman2023flow}. However, our empirical findings indicate that the matching between $x_T$ and $x_0$ in the SR task is relatively easier to learn than denoising under different noise levels, as diffusion models do. Specifically, the capacity of the student network $f_{\hat{\theta}}$ is sufficient to effectively capture the ODE process $F_{\theta}$ using only one step. To verify our assumption, we evaluate the performance of smaller models trained under different strategies. Specifically, one model is trained following the experimental settings of \cite{yue2023resshift} while the number of parameters decreases from $118.6$M to $24.3$M. Another model uses the same backbone as the aforementioned small model while directly learning the mapping relationship between $x_T$ and $\hat{x}_0$ from the standard-size teacher diffusion model. A comparison between these two small models is reported in Table~\ref{tab:ablation_small}. As demonstrated by the results, the model trained for denoising under different noise levels suffers from a serious performance drop compared with the model that directly learns the deterministic mapping between. This strongly supports our assumption that directly learning the deterministic mapping is relatively easier.

\begin{table}[h]
    \centering
    \scalebox{0.9}{
    \begin{tabular}{ccc}
    \toprule
    Methods & CLIPIQA~$\uparrow$& MUSIQ~$\uparrow$ \\
    \midrule
    ResShift~\cite{yue2023resshift} (24.32M) & 0.5365 & 52.71 \\
    ResShift~\cite{yue2023resshift} (118.59M) & \textbf{0.6537} & \textbf{61.33} \\
    \textit{SinSR} (distill only) (24.32M) & \underline{0.6499} & \underline{58.71} \\
    \bottomrule
    \end{tabular}}
    \vspace{-0.2cm}
    \caption{A comparison of the models trained with different strategies on RealSet65. The model trained with the diffusion loss, \ie, ResShift, is more sensitive to the model size than directly learning the deterministic mapping between $x_T$ and $\hat{x}_0$, indicating that the deterministic mapping is relatively easier to learn. }
    \label{tab:ablation_small}
    \vspace{-0.1cm}
\end{table}

\noindent\textbf{Is a more sophisticated distillation strategy necessary?}
To explore the necessity of more advanced techniques that learn the mapping between $x_T$ and $x_0$, we evaluate the performance of Rectified Flow~\cite{liu2022flow}, a recent method that learns the mapping to a single step through an iterative manner. Specifically, Reflow operations are conducted to avoid crossing the generation paths, and then followed by distilling the rectified generation process into a single step. However, as shown in Table~\ref{tab:compare_rectified}, the involved iterative distillation degrades the performance of the final model due to the accumulated error as discussed by the author~\cite{liu2022flow}. Besides, as verified by the previous section that the deterministic mapping between $x_T$ and $x_0$ is easy to learn in the SR task, the benefit of a more sophisticated distillation strategy is not obvious.

\begin{table}[ht]
    \centering
    \scalebox{0.9}{
    \begin{tabular}{cccc}
    \toprule
         &  LPIPS$\downarrow$ & MUSIQ$\uparrow$ & CLIPIQA$\uparrow$\\
    \midrule
    ResShift~\cite{yue2023resshift} & \underline{0.2275} & \textbf{53.90} & \underline{0.6029}\\
    w/ Rectified Flow~\cite{liu2022flow} & 0.2322 & 51.05 & 0.5753 \\
    \textit{SinSR} (distill only) & \textbf{0.2221} & 53.32 & \textbf{0.6072} \\
    \bottomrule
    \end{tabular}}
    \vspace{-0.2cm}
    \caption{A comparison between models accelerated by the proposed method and \cite{yue2023resshift}, which includes a reflow and a distillation operation. The models are evaluated on ImageNet-Test~\cite{yue2023resshift}. }
    \label{tab:compare_rectified}
\end{table}

\begin{figure}[tbp]
    \centering
    \includegraphics[width=1\linewidth]{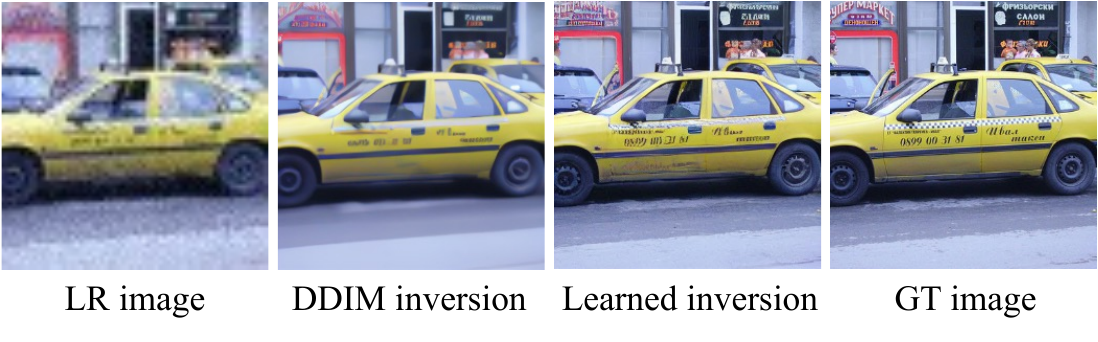}
    \vspace{-0.8cm}
    \caption{A comparison between HR images generated from DDIM inversion and the proposed learned inversion. Zoom in for details.}
    \label{fig:inversion}
    \vspace{-0.4cm}
\end{figure}

\noindent\textbf{Learned inversion.} As the core of the consistency preserving loss, a comparison with the DDIM inversion~\cite{song2020denoising} is shown in Fig.~\ref{fig:inversion}, where the proposed method achieves better fidelity performance. It indicates that the proposed method can obtain a more accurate estimation of $x_T$. Besides, more analyses regarding the consistency preserving loss are in the supplementary material.

\noindent\textbf{Training overhead.} 
While the proposed method involves solving ODEs during training, benefiting from a shortened inference process and initializing the student model from the pre-trained teacher model, the training cost of finetuning using the proposed training paradigm is still lower than that of retraining the diffusion model from scratch. Specifically, the training cost is shown in Table~\ref{tab:training_overhead}.

\vspace{-0.1cm}
\begin{table}[h]
\centering
\scalebox{0.9}{
\begin{tabular}{cccc}
\toprule
         & Num of Iters & s/Iter  & Training Time  \\
\midrule
ResShift~\cite{yue2023resshift} & 500k & 1.32s   & $\sim$7.64 days    \\
\textit{SinSR (Ours)}     & 30k & 7.41s    & $\sim$2.57 days \\ 
\bottomrule
\end{tabular}}
\vspace{-0.2cm}
\caption{A comparison of the training cost on an NVIDIA A100.}
\label{tab:training_overhead}
\vspace{-0.6cm}
\end{table}


\section{Conclusion}
In this work, we propose a novel strategy to accelerate the diffusion-based SR models into a single inference step. Specifically, a one-step bi-directional distillation is proposed to learn the deterministic mapping between the input noise and the generated high-resolution image and versa vice from a teacher diffusion model with our derived deterministic sampling. Meanwhile, a novel consistency preserving loss is optimized at the same time during the distillation so that the student model not only uses the information from the pre-trained teacher diffusion model but also directly learns from ground-truth images. Experimental results demonstrate that the proposed method can achieve on-par or even better performance than the teacher model in only one step.
{
    \small
    \bibliographystyle{ieeenat_fullname}
    \bibliography{main}

\begin{thebibliography}{47}
\providecommand{\natexlab}[1]{#1}
\providecommand{\url}[1]{\texttt{#1}}
\expandafter\ifx\csname urlstyle\endcsname\relax
  \providecommand{\doi}[1]{doi: #1}\else
  \providecommand{\doi}{doi: \begingroup \urlstyle{rm}\Url}\fi

\bibitem[Ahn et~al.(2018)Ahn, Kang, and Sohn]{ahn2018image}
Namhyuk Ahn, Byungkon Kang, and Kyung-Ah Sohn.
\newblock Image super-resolution via progressive cascading residual network.
\newblock In \emph{Proceedings of the IEEE Conference on Computer Vision and Pattern Recognition Workshops}, pages 791--799, 2018.

\bibitem[Alpher(2002)]{Alpher02}
FirstName Alpher.
\newblock Frobnication.
\newblock \emph{IEEE TPAMI}, 12\penalty0 (1):\penalty0 234--778, 2002.

\bibitem[Cai et~al.(2019)Cai, Zeng, Yong, Cao, and Zhang]{cai2019toward}
Jianrui Cai, Hui Zeng, Hongwei Yong, Zisheng Cao, and Lei Zhang.
\newblock Toward real-world single image super-resolution: A new benchmark and a new model.
\newblock In \emph{Proceedings of the IEEE/CVF International Conference on Computer Vision}, pages 3086--3095, 2019.

\bibitem[Choi et~al.(2021)Choi, Kim, Jeong, Gwon, and Yoon]{Choi_Kim_Jeong_Gwon_Yoon_2021}
Jooyoung Choi, Sungwon Kim, Yonghyun Jeong, Youngjune Gwon, and Sungroh Yoon.
\newblock Ilvr: Conditioning method for denoising diffusion probabilistic models.
\newblock In \emph{2021 IEEE/CVF International Conference on Computer Vision (ICCV)}, 2021.

\bibitem[Chung et~al.(2022)Chung, Sim, and Ye]{chung2022come}
Hyungjin Chung, Byeongsu Sim, and Jong~Chul Ye.
\newblock Come-closer-diffuse-faster: Accelerating conditional diffusion models for inverse problems through stochastic contraction.
\newblock In \emph{Proceedings of the IEEE/CVF Conference on Computer Vision and Pattern Recognition}, pages 12413--12422, 2022.

\bibitem[Dahl et~al.(2017)Dahl, Norouzi, and Shlens]{Dahl_Norouzi_Shlens_2017}
Ryan Dahl, Mohammad Norouzi, and Jonathon Shlens.
\newblock Pixel recursive super resolution.
\newblock In \emph{2017 IEEE International Conference on Computer Vision (ICCV)}, 2017.

\bibitem[Deng et~al.(2009)Deng, Dong, Socher, Li, Li, and Fei-Fei]{deng2009imagenet}
Jia Deng, Wei Dong, Richard Socher, Li-Jia Li, Kai Li, and Li Fei-Fei.
\newblock Imagenet: A large-scale hierarchical image database.
\newblock In \emph{2009 IEEE conference on computer vision and pattern recognition}, pages 248--255. Ieee, 2009.

\bibitem[Dong et~al.(2015)Dong, Loy, He, and Tang]{dong2015image}
Chao Dong, Chen~Change Loy, Kaiming He, and Xiaoou Tang.
\newblock Image super-resolution using deep convolutional networks.
\newblock \emph{IEEE transactions on pattern analysis and machine intelligence}, 38\penalty0 (2):\penalty0 295--307, 2015.

\bibitem[Guo et~al.(2022)Guo, Zhang, Wu, Wang, Zhang, and Wang]{guo2022lar}
Baisong Guo, Xiaoyun Zhang, Haoning Wu, Yu Wang, Ya Zhang, and Yan-Feng Wang.
\newblock Lar-sr: A local autoregressive model for image super-resolution.
\newblock In \emph{Proceedings of the IEEE/CVF Conference on Computer Vision and Pattern Recognition}, pages 1909--1918, 2022.

\bibitem[Ho et~al.(2020)Ho, Jain, and Abbeel]{ddpm}
Jonathan Ho, Ajay Jain, and Pieter Abbeel.
\newblock Denoising diffusion probabilistic models.
\newblock In \emph{Advances in Neural Information Processing Systems}, pages 6840--6851. Curran Associates, Inc., 2020.

\bibitem[Ji et~al.(2020)Ji, Cao, Tai, Wang, Li, and Huang]{ji2020real}
Xiaozhong Ji, Yun Cao, Ying Tai, Chengjie Wang, Jilin Li, and Feiyue Huang.
\newblock Real-world super-resolution via kernel estimation and noise injection.
\newblock In \emph{proceedings of the IEEE/CVF conference on computer vision and pattern recognition workshops}, pages 466--467, 2020.

\bibitem[Karras et~al.(2018)Karras, Aila, Laine, and Lehtinen]{karras2018progressive}
Tero Karras, Timo Aila, Samuli Laine, and Jaakko Lehtinen.
\newblock Progressive growing of gans for improved quality, stability, and variation.
\newblock In \emph{International Conference on Learning Representations}, 2018.

\bibitem[Kawar et~al.(2022)Kawar, Elad, Ermon, and Song]{kawar2022denoising}
Bahjat Kawar, Michael Elad, Stefano Ermon, and Jiaming Song.
\newblock Denoising diffusion restoration models.
\newblock \emph{Advances in Neural Information Processing Systems}, 35:\penalty0 23593--23606, 2022.

\bibitem[Ke et~al.(2021)Ke, Wang, Wang, Milanfar, and Yang]{ke2021musiq}
Junjie Ke, Qifei Wang, Yilin Wang, Peyman Milanfar, and Feng Yang.
\newblock Musiq: Multi-scale image quality transformer.
\newblock In \emph{Proceedings of the IEEE/CVF International Conference on Computer Vision}, pages 5148--5157, 2021.

\bibitem[Kim et~al.(2016)Kim, Lee, and Lee]{Kim_Lee_Lee_2016}
Jiwon Kim, Jung~Kwon Lee, and Kyoung~Mu Lee.
\newblock Accurate image super-resolution using very deep convolutional networks.
\newblock In \emph{2016 IEEE Conference on Computer Vision and Pattern Recognition (CVPR)}, 2016.

\bibitem[Ledig et~al.(2017)Ledig, Theis, Husz{\'a}r, Caballero, Cunningham, Acosta, Aitken, Tejani, Totz, Wang, et~al.]{ledig2017photo}
Christian Ledig, Lucas Theis, Ferenc Husz{\'a}r, Jose Caballero, Andrew Cunningham, Alejandro Acosta, Andrew Aitken, Alykhan Tejani, Johannes Totz, Zehan Wang, et~al.
\newblock Photo-realistic single image super-resolution using a generative adversarial network.
\newblock In \emph{Proceedings of the IEEE conference on computer vision and pattern recognition}, pages 4681--4690, 2017.

\bibitem[Liang et~al.(2021)Liang, Cao, Sun, Zhang, Van~Gool, and Timofte]{liang2021swinir}
Jingyun Liang, Jiezhang Cao, Guolei Sun, Kai Zhang, Luc Van~Gool, and Radu Timofte.
\newblock Swinir: Image restoration using swin transformer.
\newblock In \emph{Proceedings of the IEEE/CVF international conference on computer vision}, pages 1833--1844, 2021.

\bibitem[Liang et~al.(2022)Liang, Zeng, and Zhang]{liang2022efficient}
Jie Liang, Hui Zeng, and Lei Zhang.
\newblock Efficient and degradation-adaptive network for real-world image super-resolution.
\newblock In \emph{European Conference on Computer Vision}, pages 574--591. Springer, 2022.

\bibitem[Lipman et~al.(2023)Lipman, Chen, Ben-Hamu, Nickel, and Le]{lipman2023flow}
Yaron Lipman, Ricky T.~Q. Chen, Heli Ben-Hamu, Maximilian Nickel, and Matthew Le.
\newblock Flow matching for generative modeling.
\newblock In \emph{The Eleventh International Conference on Learning Representations}, 2023.

\bibitem[Liu et~al.(2022)Liu, Gong, et~al.]{liu2022flow}
Xingchao Liu, Chengyue Gong, et~al.
\newblock Flow straight and fast: Learning to generate and transfer data with rectified flow.
\newblock In \emph{The Eleventh International Conference on Learning Representations}, 2022.

\bibitem[Lu et~al.(2022)Lu, Zhou, Bao, Chen, Li, and Zhu]{lu2022dpm}
Cheng Lu, Yuhao Zhou, Fan Bao, Jianfei Chen, Chongxuan Li, and Jun Zhu.
\newblock Dpm-solver: A fast ode solver for diffusion probabilistic model sampling in around 10 steps.
\newblock \emph{Advances in Neural Information Processing Systems}, 35:\penalty0 5775--5787, 2022.

\bibitem[Lugmayr et~al.(2020)Lugmayr, Danelljan, Van~Gool, and Timofte]{lugmayr2020srflow}
Andreas Lugmayr, Martin Danelljan, Luc Van~Gool, and Radu Timofte.
\newblock Srflow: Learning the super-resolution space with normalizing flow.
\newblock In \emph{Computer Vision--ECCV 2020: 16th European Conference, Glasgow, UK, August 23--28, 2020, Proceedings, Part V 16}, pages 715--732. Springer, 2020.

\bibitem[Luhman and Luhman(2021)]{luhman2021knowledge}
Eric Luhman and Troy Luhman.
\newblock Knowledge distillation in iterative generative models for improved sampling speed.
\newblock \emph{arXiv preprint arXiv:2101.02388}, 2021.

\bibitem[Meng et~al.(2023)Meng, Rombach, Gao, Kingma, Ermon, Ho, and Salimans]{meng2023distillation}
Chenlin Meng, Robin Rombach, Ruiqi Gao, Diederik Kingma, Stefano Ermon, Jonathan Ho, and Tim Salimans.
\newblock On distillation of guided diffusion models.
\newblock In \emph{Proceedings of the IEEE/CVF Conference on Computer Vision and Pattern Recognition}, pages 14297--14306, 2023.

\bibitem[Menick and Kalchbrenner(2018)]{Menick_Kalchbrenner_2018}
Jacob Menick and Nal Kalchbrenner.
\newblock Generating high fidelity images with subscale pixel networks and multidimensional upscaling.
\newblock \emph{International Conference on Learning Representations,International Conference on Learning Representations}, 2018.

\bibitem[Menon et~al.(2020)Menon, Damian, Hu, Ravi, and Rudin]{Menon_Damian_Hu_Ravi_Rudin_2020}
Sachit Menon, Alexandru Damian, Shijia Hu, Nikhil Ravi, and Cynthia Rudin.
\newblock Pulse: Self-supervised photo upsampling via latent space exploration of generative models.
\newblock In \emph{2020 IEEE/CVF Conference on Computer Vision and Pattern Recognition (CVPR)}, 2020.

\bibitem[Nichol and Dhariwal(2021)]{nichol2021improved}
Alexander~Quinn Nichol and Prafulla Dhariwal.
\newblock Improved denoising diffusion probabilistic models.
\newblock In \emph{International Conference on Machine Learning}, pages 8162--8171. PMLR, 2021.

\bibitem[Oord et~al.(2016)Oord, Kalchbrenner, Vinyals, Espeholt, Graves, and Kavukcuoglu]{Oord_Kalchbrenner_Vinyals_Espeholt_Graves_Kavukcuoglu_2016}
Aaronvanden Oord, Nal Kalchbrenner, Oriol Vinyals, Lasse Espeholt, Alex Graves, and Koray Kavukcuoglu.
\newblock Conditional image generation with pixelcnn decoders.
\newblock \emph{arXiv: Computer Vision and Pattern Recognition,arXiv: Computer Vision and Pattern Recognition}, 2016.

\bibitem[Parmar et~al.(2018)Parmar, Vaswani, Uszkoreit, Kaiser, Shazeer, Ku, and Tran]{Parmar_Vaswani_Uszkoreit_Kaiser_Shazeer_Ku_Tran_2018}
Niki Parmar, Ashish Vaswani, Jakob Uszkoreit, Łukasz Kaiser, Noam Shazeer, Alexander Ku, and Dustin Tran.
\newblock Image transformer.
\newblock \emph{arXiv: Computer Vision and Pattern Recognition,arXiv: Computer Vision and Pattern Recognition}, 2018.

\bibitem[Radford et~al.(2021)Radford, Kim, Hallacy, Ramesh, Goh, Agarwal, Sastry, Askell, Mishkin, Clark, et~al.]{radford2021learning}
Alec Radford, Jong~Wook Kim, Chris Hallacy, Aditya Ramesh, Gabriel Goh, Sandhini Agarwal, Girish Sastry, Amanda Askell, Pamela Mishkin, Jack Clark, et~al.
\newblock Learning transferable visual models from natural language supervision.
\newblock In \emph{International conference on machine learning}, pages 8748--8763. PMLR, 2021.

\bibitem[Rombach et~al.(2022)Rombach, Blattmann, Lorenz, Esser, and Ommer]{rombach2022high}
Robin Rombach, Andreas Blattmann, Dominik Lorenz, Patrick Esser, and Bj{\"o}rn Ommer.
\newblock High-resolution image synthesis with latent diffusion models.
\newblock In \emph{Proceedings of the IEEE/CVF conference on computer vision and pattern recognition}, pages 10684--10695, 2022.

\bibitem[Saharia et~al.(2022)Saharia, Ho, Chan, Salimans, Fleet, and Norouzi]{saharia2022image}
Chitwan Saharia, Jonathan Ho, William Chan, Tim Salimans, David~J Fleet, and Mohammad Norouzi.
\newblock Image super-resolution via iterative refinement.
\newblock \emph{IEEE Transactions on Pattern Analysis and Machine Intelligence}, 45\penalty0 (4):\penalty0 4713--4726, 2022.

\bibitem[Sajjadi et~al.(2017)Sajjadi, Scholkopf, and Hirsch]{Sajjadi_Scholkopf_Hirsch_2017}
Mehdi S.~M. Sajjadi, Bernhard Scholkopf, and Michael Hirsch.
\newblock Enhancenet: Single image super-resolution through automated texture synthesis.
\newblock In \emph{2017 IEEE International Conference on Computer Vision (ICCV)}, 2017.

\bibitem[Salimans and Ho(2021)]{salimans2021progressive}
Tim Salimans and Jonathan Ho.
\newblock Progressive distillation for fast sampling of diffusion models.
\newblock In \emph{International Conference on Learning Representations}, 2021.

\bibitem[Schuhmann et~al.(2021)Schuhmann, Vencu, Beaumont, Kaczmarczyk, Mullis, Katta, Coombes, Jitsev, and Komatsuzaki]{schuhmann2021laion}
Christoph Schuhmann, Richard Vencu, Romain Beaumont, Robert Kaczmarczyk, Clayton Mullis, Aarush Katta, Theo Coombes, Jenia Jitsev, and Aran Komatsuzaki.
\newblock Laion-400m: Open dataset of clip-filtered 400 million image-text pairs.
\newblock \emph{arXiv preprint arXiv:2111.02114}, 2021.

\bibitem[Song et~al.(2020)Song, Meng, and Ermon]{song2020denoising}
Jiaming Song, Chenlin Meng, and Stefano Ermon.
\newblock Denoising diffusion implicit models.
\newblock In \emph{International Conference on Learning Representations}, 2020.

\bibitem[Song et~al.(2023)Song, Dhariwal, Chen, and Sutskever]{song2023consistency}
Yang Song, Prafulla Dhariwal, Mark Chen, and Ilya Sutskever.
\newblock Consistency models.
\newblock 2023.

\bibitem[Wang et~al.(2023{\natexlab{a}})Wang, Chan, and Loy]{wang2023exploring}
Jianyi Wang, Kelvin~CK Chan, and Chen~Change Loy.
\newblock Exploring clip for assessing the look and feel of images.
\newblock In \emph{Proceedings of the AAAI Conference on Artificial Intelligence}, pages 2555--2563, 2023{\natexlab{a}}.

\bibitem[Wang et~al.(2018)Wang, Yu, Wu, Gu, Liu, Dong, Qiao, and Change~Loy]{wang2018esrgan}
Xintao Wang, Ke Yu, Shixiang Wu, Jinjin Gu, Yihao Liu, Chao Dong, Yu Qiao, and Chen Change~Loy.
\newblock Esrgan: Enhanced super-resolution generative adversarial networks.
\newblock In \emph{Proceedings of the European conference on computer vision (ECCV) workshops}, pages 0--0, 2018.

\bibitem[Wang et~al.(2021)Wang, Xie, Dong, and Shan]{wang2021real}
Xintao Wang, Liangbin Xie, Chao Dong, and Ying Shan.
\newblock Real-esrgan: Training real-world blind super-resolution with pure synthetic data.
\newblock In \emph{Proceedings of the IEEE/CVF international conference on computer vision}, pages 1905--1914, 2021.

\bibitem[Wang et~al.(2022)Wang, Wan, Yang, Li, Chau, and Kot]{wang2022low}
Yufei Wang, Renjie Wan, Wenhan Yang, Haoliang Li, Lap-Pui Chau, and Alex Kot.
\newblock Low-light image enhancement with normalizing flow.
\newblock In \emph{Proceedings of the AAAI conference on artificial intelligence}, pages 2604--2612, 2022.

\bibitem[Wang et~al.(2023{\natexlab{b}})Wang, Yu, Yang, Guo, Chau, Kot, and Wen]{wang2023exposurediffusion}
Yufei Wang, Yi Yu, Wenhan Yang, Lanqing Guo, Lap-Pui Chau, Alex~C Kot, and Bihan Wen.
\newblock Exposurediffusion: Learning to expose for low-light image enhancement.
\newblock In \emph{Proceedings of the IEEE/CVF International Conference on Computer Vision}, pages 12438--12448, 2023{\natexlab{b}}.

\bibitem[Wang et~al.(2015)Wang, Liu, Yang, Han, and Huang]{Wang_Liu_Yang_Han_Huang_2015}
Zhaowen Wang, Ding Liu, Jianchao Yang, Wei Han, and ThomasS. Huang.
\newblock Deep networks for image super-resolution with sparse prior.
\newblock \emph{Cornell University - arXiv,Cornell University - arXiv}, 2015.

\bibitem[Wang et~al.(2020)Wang, Chen, and Hoi]{wang2020deep}
Zhihao Wang, Jian Chen, and Steven~CH Hoi.
\newblock Deep learning for image super-resolution: A survey.
\newblock \emph{IEEE transactions on pattern analysis and machine intelligence}, 43\penalty0 (10):\penalty0 3365--3387, 2020.

\bibitem[Yue et~al.(2023)Yue, Wang, and Loy]{yue2023resshift}
Zongsheng Yue, Jianyi Wang, and Chen~Change Loy.
\newblock Resshift: Efficient diffusion model for image super-resolution by residual shifting.
\newblock \emph{Advances in Neural Information Processing Systems}, 2023.

\bibitem[Zhang et~al.(2021)Zhang, Liang, Van~Gool, and Timofte]{zhang2021designing}
Kai Zhang, Jingyun Liang, Luc Van~Gool, and Radu Timofte.
\newblock Designing a practical degradation model for deep blind image super-resolution.
\newblock In \emph{Proceedings of the IEEE/CVF International Conference on Computer Vision}, pages 4791--4800, 2021.

\bibitem[Zhang et~al.(2018)Zhang, Isola, Efros, Shechtman, and Wang]{zhang2018unreasonable}
Richard Zhang, Phillip Isola, Alexei~A Efros, Eli Shechtman, and Oliver Wang.
\newblock The unreasonable effectiveness of deep features as a perceptual metric.
\newblock In \emph{Proceedings of the IEEE conference on computer vision and pattern recognition}, pages 586--595, 2018.

\end{thebibliography}
}


\end{document}